\documentclass[runningheads]{llncs}

\usepackage{abbrv}
\usepackage[a4paper, total={6in, 8in}]{geometry}

\usepackage{graphicx}
\usepackage{subcaption}
\usepackage{booktabs}

\usepackage{amsmath}
\usepackage{amssymb}
\usepackage[ruled,vlined,linesnumbered]{algorithm2e}

\usepackage[accsupp]{axessibility}  

\usepackage{hyperref}

\usepackage{orcidlink}

\usepackage[capitalize]{cleveref}

\begin{document}

\title{NeoNeXt: Novel neural network operator and architecture based on the patch-wise matrix multiplications} 

\titlerunning{NeoNeXt}

\author{Vladimir Korviakov\inst{1}\orcidlink{0009-0000-5019-5633} \and
Denis Koposov\inst{1}\orcidlink{0009-0004-1840-9637}}

\authorrunning{V.~Korviakov, D.~Koposov}

\institute{Moscow Media Technology lab, Huawei Technologies Co., Ltd
\email{{korviakov.vladimir1,koposov.denis1}@huawei.com}\\
\url{https://www.huawei.com/}}

\maketitle

\begin{abstract}
   Most of the computer vision architectures nowadays are built upon the well-known foundation operations: fully-connected layers, convolutions and multi-head self-attention blocks. In this paper we propose a novel foundation operation - NeoCell - which learns matrix patterns and performs patchwise matrix multiplications with the input data. The main advantages of the proposed operator are (1) simple implementation without need in operations like im2col, (2) low computational complexity (especially for large matrices) and (3) simple and flexible implementation of up-/down-sampling. We validate NeoNeXt family of models based on this operation on ImageNet-1K classification task and show that they achieve competitive quality.
  \keywords{Neural Network architecture \and Computer Vision}
\end{abstract}

\section{Introduction}
A plethora of the specialized AI hardware such as Neural Processing Units (NPUs) have appeared recently.
These devices are typically good at running parallelizable tasks of tensor and matrix multiplications and additions, however they also set specific limitations for the models to be deployed.
For example, NPUs provide fast vector instructions and very efficient large matrix multiplications, but the memory operations are relatively slow.
Non-vectorizable operations or dynamic model structure should also be avoided.

Thus, it is important to explore new basic paradigms of the Neural Networks, which can utilize the advantages of NPUs and avoid their disadvantages, while keeping the quality of the solution of the target problem.

In this paper we continue this exploration by considering Huawei Ascend~\cite{huawei2024ascend} based on Da Vinci architecture~\cite{liao2019davinci} as our target platform.
This device has been designed to maximize the efficiency of the most popular neural architectures at the moment of design, such as ResNet~\cite{He2016}.
These architectures are based on the convolution operation that can be computed in several ways.
The most popular algorithm is to transform the input data of shape $C_{in}\times H_{in} \times W_{in}$ to the matrix of size $H_{in} W_{in} \times C_{in} K^2$ using \textit{im2col} operation~\cite{Chellapilla2006HighPC}.
This matrix is then multiplied by the matrix of shape $C_{in} K^2 \times C_{out}$ obtained from the convolutional weights by reshaping them.
Practically, im2col operation increases the size of input data by $K^2$ times, where K is the size of kernel (square kernel is considered for simplicity).

To solve the aforementioned problems of the convolution we propose a novel operation - \textbf{NeoCell} - that is built upon a sequence of matrix multiplications in a way that differs from all known architectures existing today.
We also construct a family of architectures called \textbf{NeoNeXt} based on the ConvNeXt~\cite{liu2022convnet}, while using this newly designed operation instead of 2D convolutional layers.
NeoCell essentially multiplies the input by two matrices of weights from the left side and from the right side.

The most straightforward application for this operator is to replace the depthwise convolution, which is inferior in several aspects.
First of all, NeoCell is $2/K$ times less complex (see \cref{complexity}), where $K$ is the convolutional kernel size and NeoCell matrix size.
Secondly, it does not require im2col and allows to reduce the amount of memory required by the intermediate representations.
NeoCell also allows to do flexible data subsampling with a fractional scaling factor, which is described in \cref{subsampling}.
Finally, implementation of NeoCell is essentially a series of two multiplications by the large structured matrices, which is very efficient and can be done on any device.

We test our family of the models on the ImageNet classification task and achieve the performance competitive to the modern architectures. The details of the experiments are provided in \cref{imagenet_experiments}.

\begin{figure}[tb]
\begin{center}
\includegraphics[width=1.0\linewidth]{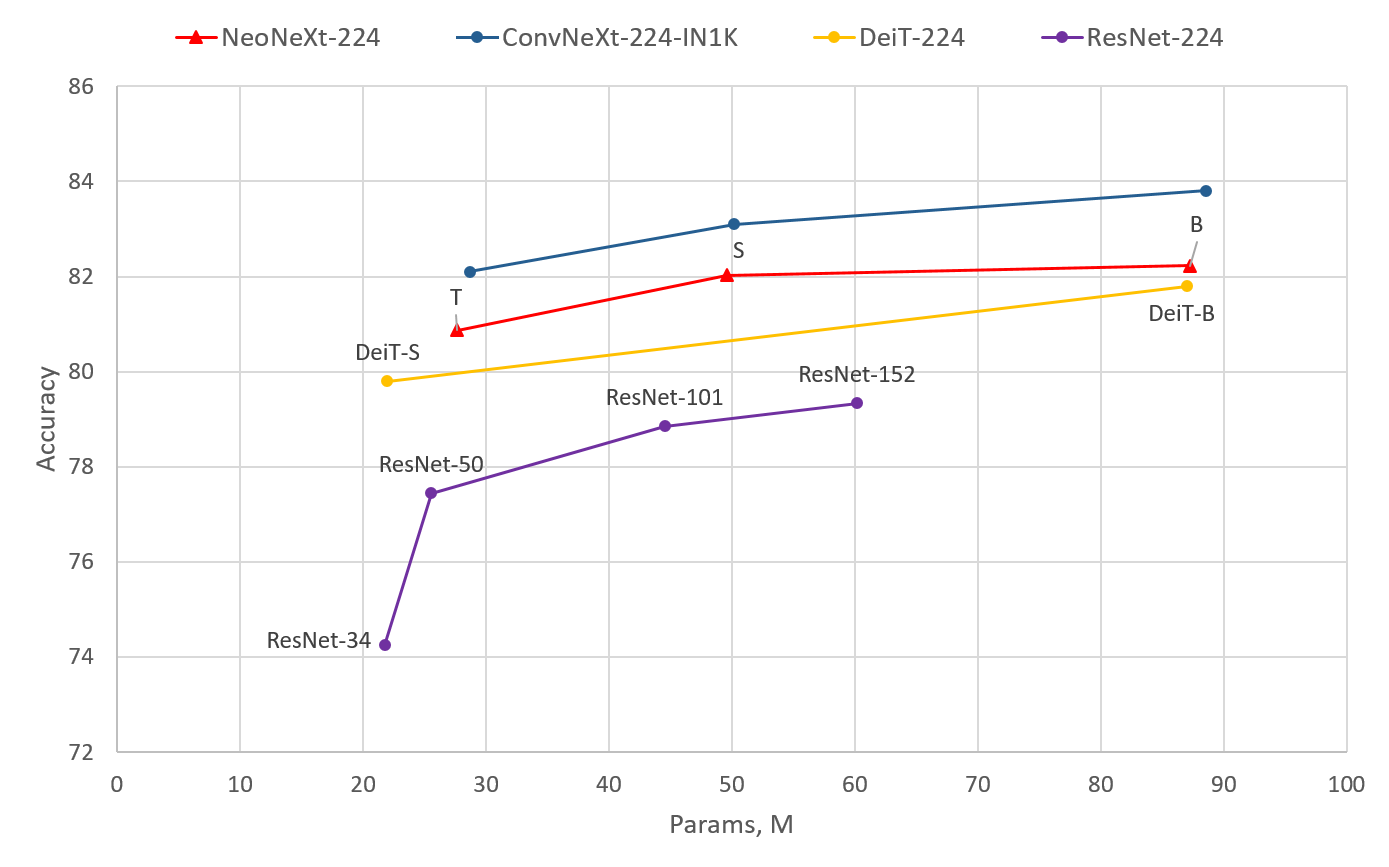}
\end{center}
\caption{\textbf{ImageNet-1K accuracy vs parameters}. NeoNeXt is compared to ResNet, DeiT and ConvNeXt. Alough we do not achieve state-of-the-art performance, we prove that NeoCell operation is a viable choice. It's also shown that NeoNeXt quality grows with the number of parameters, which means that models are scalable. \textbf{IN1K} means that ConvNeXt was not pretrained on external data such as ImageNet-22K.}
\label{fig:neo_acc_params}
\end{figure}

\section{Related Work}
A lot of vendors have recently addressed the problem of AI cost minimization by designing specialized AI acceleration hardware and platforms.
The examples of such platforms are Google Cloud TPU~\cite{tpu2019edge,10.1145/3154484}, Nvidia Jetson~\cite{nvidiajetson}, Huawei Ascend~\cite{huawei2024ascend}, Intel Movidius Myriad~\cite{intelmovidiusmyriad} and etc.

Typical way to minimize the latency and maximize the throughput is to design specialized low-level kernels for existing operators.
FlashAttention~\cite{NEURIPS2022_67d57c32} and FlashAttention-2~\cite{dao2023flashattention2} are good examples of such operators.

The classical convolutional operator optimizations are based on Winograd method~\cite{doi:10.1137/1.9781611970364,Lavin2015FastAF}, im2col method~\cite{Chellapilla2006HighPC} or Fast Fourier Transform and Convolution Theorem~\cite{Mathieu2013FastTO}.

Algorithmic optimizations can have one big benefit: they can be exactly or approximately equivalent to non-optimized methods, thus existing pretrained models can be accelerated without training from scratch.
The disadvantage of algorithmic optimizations is that the designed operators are limited by the theoretical properties of the baseline method.
Thus it is important to explore novel fundamental algorithms that allow to outperform existing methods.

One of the previous attempts to optimize convolutions for specialized devices was depthwise separable convolution mostly known from the MobileNet and EfficientNet families of models.
The usual convolution is replaced by two convolutional layers: spatial (acts on each channel separately) and pointwise (has a kernel of size 1).
It allows to reduce the amount of memory consumed by the model parameters and make it less computationally complex without any effects on the quality.
However, the problem of intermediate representation size during im2col stays.

One more possible way to increase the efficiency of the hardware utilization is to design hardware-friendly neural architectures using Neural Architecture Search methods~\cite{tan2019mnasnet,letunovskiy2022isynet,BAYMURZINA202282}.
The disadvantage of NAS-based methods is that only limited set of already designed operators can be included into the search space.

Another notable example of the novel architecture paradigm is MLP-Mixer architecture~\cite{NEURIPS2021_cba0a4ee} that builds a competitive model based exclusively on multi-layer perceptrons (MLPs).

\section{Method}
\subsection{NeoCell Operator}
\begin{figure*}
\begin{center}
\includegraphics[width=1.0\linewidth]{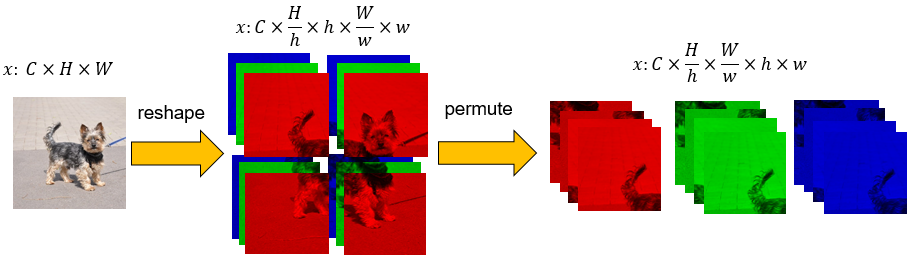}
\end{center}
   \caption{Illustration of splitting of the data into the patches}
\label{fig:patches}
\end{figure*}

The key idea of the proposed operator ("NeoCell") is the following: given the input data tensor $X$ of shape $[C \times H \times W]$ (batch dimension is omitted for simplicity) we define two trainable weight tensors: $L$ of shape $[C \times h' \times h]$ and $R$ of shape $[C \times w \times w']$.
These tensors can be considered as $C$ matrices of shape $[h' \times h]$ and $[w \times w']$ correspondingly.
Then input can be reshaped to shape $[C \times \dfrac{H}{h} \times h \times \dfrac{W}{w} \times w]$ and permuted to shape $[C \times \dfrac{H}{h} \times \dfrac{W}{w} \times h \times w]$ (\cref{fig:patches}).
Then the output can be computed as
\begin{equation}
\begin{gathered}
Y_{c,i,j} = L_{c}X_{c,i,j}R_{c}, \\
c \in [1,..,C], i \in [1,..,\dfrac{H}{h}], j \in [1,..,\dfrac{W}{w}]
\end{gathered}
\end{equation}
In other words, for each channel $c$ we split data into $\dfrac{H}{h} \cdot \dfrac{W}{w}$ patches of shape $[h \times w]$ and each of them multiplied by matrices $L_{c}$ (left side multiplication) and $R_{c}$ (right side multiplication).
Optionally, trainable bias $B$ of shape $[C \times h' \times w']$ can be added to the result.
The output has shape $[C \times \dfrac{H}{h} \times \dfrac{W}{w} \times h' \times w']$ which is permuted to $[C \times \dfrac{H}{h} \times h' \times \dfrac{W}{w} \times w']$ and reshaped to $[C \times \dfrac{H}{h} \cdot h' \times \dfrac{W}{w} \cdot w']$.

The proposed operation together with the other common operations like activation functions, normalization, fully-connected layers (or pointwise convolutions) and skip connection allows to construct the neural of the stage structure: down-sampling of the data is allowed between the stages, while within each stage spatial shape of the data is not changed. In \cref{block_diagonal} we show mathematically equivalent formulation of the NeoCell operator without need in reshape and permute operations.

\subsection{Up- and Down-sampling} \label{subsampling}

\begin{figure}[tb]
  \centering
  \begin{subfigure}{1.0\linewidth}
    \includegraphics[width=1.0\linewidth]{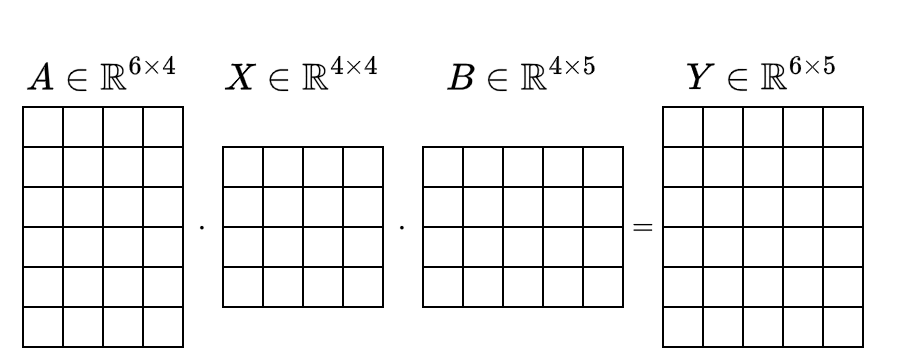}
    \caption{Using NeoCell operator for up-sampling}
    \label{fig:upsampling}
  \end{subfigure}
  \vfill
  \begin{subfigure}{1.0\linewidth}
    \includegraphics[width=1.0\linewidth]{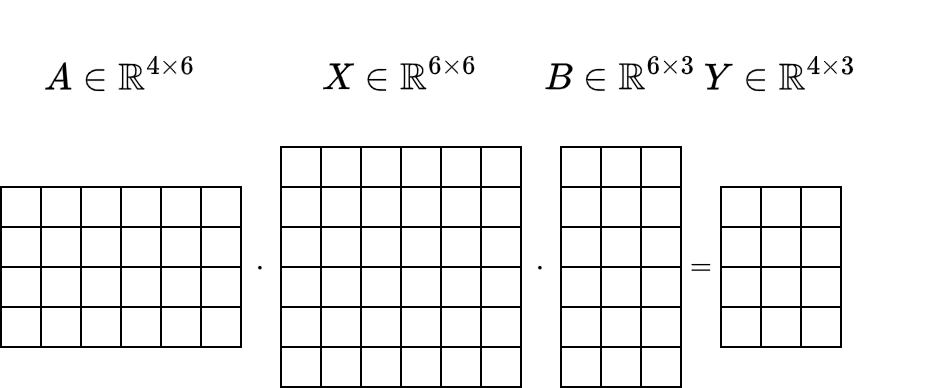}
    \caption{Using NeoCell operator for down-sampling}
    \label{fig:downsampling}
  \end{subfigure}
  \caption{Methods of spatial size change}
  \label{fig:spatial_change}
\end{figure}

Obviously, if weight matrices are square, i.e. $h'=h$ and $w' = w$, then the shape of input and output data are the same.
If $h' < h$ and/or $w' < w$ then NeoCell performs down-sampling of the input data by one or both of the spatial dimensions.
If $h' > h$ and/or $w' > w$ then NeoCell performs up-sampling of the input data by one or both of the spatial dimensions.
This property enables flexible change of the data spatial size.
The simple illustrations of up- and down-sampling are shown on \cref{fig:upsampling} and \cref{fig:downsampling} correspondingly.

\subsection{Inter-channel Information Exchange}

Conceptually, NeoCell can be considered as a replacement of depthwise convolution.
It means that single NeoCell operation connects each $c$-th input channel to the corresponding output channel, and there is no information exchange between different channels.
To enable the information exchange between the channels we utilize pointwise convolutions (or fully-connected layers applied to the channel dimension).
This approach used in a number of modern architectures like MobileNet~\cite{Howard2017} and EfficientNet~\cite{Tan2019efficientnet}.

\subsection{Block-diagonal Implementation}\label{block_diagonal}
In practice it may be inefficient to permute the data and multiply matrices of small size.
Modern device like Huawei Ascend~\cite{huawei2024ascend} operate with larger matrices with higher efficiency.
However, it may be hard to train model with large weight filters and special tricks should be applied~\cite{ding2022replknet,liu2022}.

To overcome this contradiction we propose to reformulate the NeoCell operation in terms of the block-diagonal matrix multiplication.
In this case, the data $X$ is kept in the original format $[C \times H \times W]$ but interpreted as block matrix, where blocks $X_{ij}$ correspond to the patches in the original formulation.
Left and right matrices $A_c$ and $B_c$ of channel $c \in [1..C]$ are constructed as block-diagonal matrices with the original matrices used as the shared blocks.

\begin{equation}
\begin{gathered}
\mathbf{A_c} = \begin{bmatrix}
  \mathbf{L_c} & \mathbf{0} & \cdots & \mathbf{0} \\
  \mathbf{0} & \mathbf{L_c} &\cdots & \mathbf{0} \\
  \vdots          & \vdots          & \ddots & \vdots \\
  \mathbf{0} &\cdots & \mathbf{0} & \mathbf{L_c} \\
\end{bmatrix};\quad \mathbf{B_c} = \begin{bmatrix}
  \mathbf{R_c} & \mathbf{0} & \cdots & \mathbf{0} \\
  \mathbf{0} & \mathbf{R_c} &\cdots & \mathbf{0} \\
  \vdots          & \vdots          & \ddots & \vdots \\
  \mathbf{0} &\cdots & \mathbf{0} & \mathbf{R_c} \\
\end{bmatrix}
\end{gathered}
\end{equation}

Now the output data can be computed as $Y_c = A_{c}X_{c}B_{c}$ for each $c \in [1..C]$. The simplest example of this idea is shown in the \cref{eq:block_diagonal}. The input data is split into four virtual patches, without actual change of the shape. This formulation allows to better utilize AI acceleration devices. In our experiments this idea allowed to accelerate the implementation 15 times on NVidia V100 GPU and PyTorch framework. Additional acceleration can be achieved using specialized low-level sparse matrix multiplication operators.

\begin{equation}\label{eq:block_diagonal}
\begin{aligned}[b]
Y_c = A_{c}X_{c}B_{c} = 
\begin{bmatrix}
  \mathbf{L_c} & \mathbf{0} \\
  \mathbf{0} & \mathbf{L_c} \\
\end{bmatrix}
\begin{bmatrix}
  \mathbf{X_{c11}} & \mathbf{X_{c12}} \\
  \mathbf{X_{c21}} & \mathbf{X_{c22}} \\
\end{bmatrix}
\begin{bmatrix}
  \mathbf{R_c} & \mathbf{0} \\
  \mathbf{0} & \mathbf{R_c} \\
\end{bmatrix}
= \begin{bmatrix}
  \mathbf{{L_c}X_{c11}{R_c}} & \mathbf{{L_c}X_{c12}{R_c}} \\
  \mathbf{{L_c}X_{c21}{R_c}} & \mathbf{{L_c}X_{c22}{R_c}} \\
\end{bmatrix}
\end{aligned}
\end{equation}

\subsection{Inter-patch Information Exchange} \label{interpatch}
\begin{figure}
    \centering
    \begin{subfigure}[b]{0.45\textwidth}
        \includegraphics[width=\textwidth]{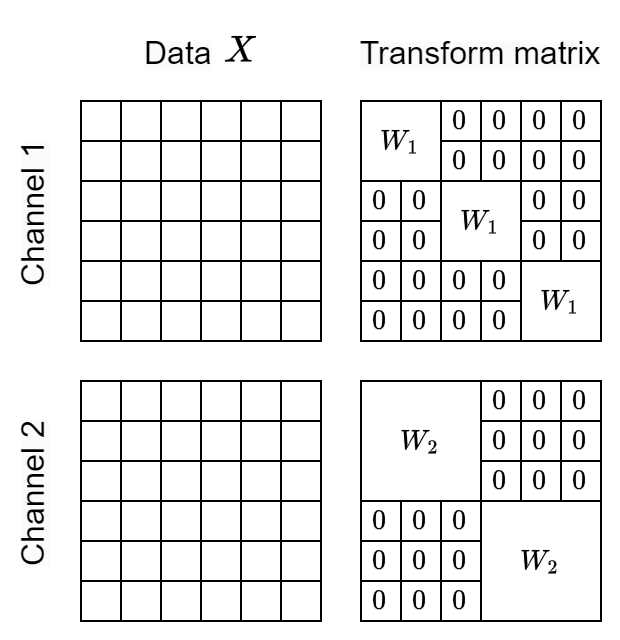}
        \caption{Using different matrix sizes for different channels}
        \label{fig:multi_kernel}
    \end{subfigure}
    \hfill
    \begin{subfigure}[b]{0.45\textwidth}
        \includegraphics[width=\textwidth]{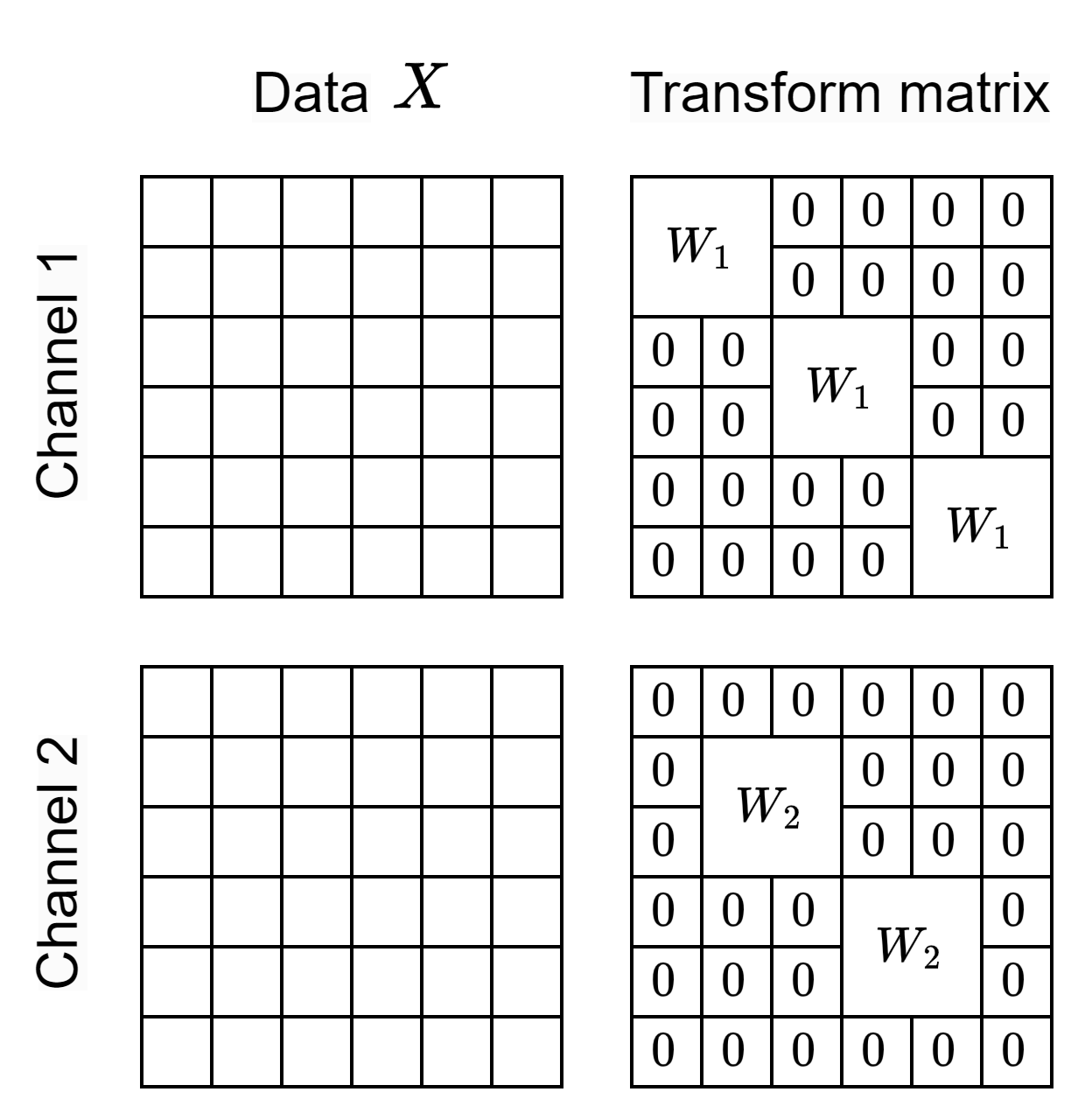}
        \caption{Using spatially shifted matrices for for different channels}
        \label{fig:spatial_shift}
    \end{subfigure}
    \caption{Methods of inter-patch information exchange. Only right NeoCell matrices are shown here.}
\end{figure}

Unlike regular convolutions, within a stage NeoCell has constant receptive field which is equal to the patch size, even for a single layer.
The receptive field does not increase for the deeper layers.
Pointwise convolutions do not solve this problem, because they also do not increase the receptive field.
To address this problem two different solutions are proposed:
\begin{enumerate}
    \item Construct block-diagonal NeoCell with different matrix sizes for different channels (\cref{fig:multi_kernel}). For example, first group of channels uses matrix 2x2, second group – 3x3, etc.
    \item Construct block-diagonal NeoCell with spatially shifted matrices for different channels (\cref{fig:spatial_shift}). For example, all channels use matrix 3x3, and in the first group of channels matrices are not shifted, in the second group – shifted by 1 pixel by width and height dimension, in the third group – shifted by 2 pixels. Instead of zeroing, the upper-left and lower-right corners of block-diagonal matrix can be filled with corresponding lower-right and upper-left parts of the weight matrix.
\end{enumerate}

These ideas can be combined together: all channels can be split into several groups with different matrix sizes, and each group can be split into $n_i$ subgroups with incrementally growing spatial shift ($n_i$ is matrix size for the $i$-th group).
Together with the pointwise convolutions proposed ideas lead to information exchange between the adjacent regions of the data, thus receptive field grows with the depth of the network.

\subsection{Complexity analysis} \label{complexity}
Consider input tensor of shape $[C \times H \times W]$, depthwise convolution with kernel $[k \times k]$ and NeoCell with both left and right weight matrices of size $[k \times k]$ (no bias in both cases). Then, the complexity of depthwise convolution (in terms of number of multiplication operations) is
\begin{equation}
\begin{gathered}
T_{DWC} = C \cdot H \cdot W \cdot k^2
\end{gathered}
\end{equation}
Assuming the na\"ive matrix multiplication algorithm with $\mathcal{O}(n^3)$ complexity, the complexity of NeoCell is
\begin{equation}\label{eq:neocell_complexity}
\begin{gathered}
T_{NEO} = 2 \cdot C \cdot \dfrac{H}{k} \cdot \dfrac{W}{k} \cdot k^3 = 2 \cdot C \cdot H \cdot W \cdot k
\end{gathered}
\end{equation}
Thus $T_{NEO} < T_{DWC}$ if $k > 2$.

As an additional advantage, NeoCell operator doesn't need extra memory operation on the input data such as im2col~\cite{Chellapilla2006HighPC} which may be important in certain cases.

\subsection{Initialization method} \label{neoinit}

We have found that proper initialization method is very important for the final quality of the model. We use \cref{alg:neo_init} to initialize each NeoCell matrix.

\SetKwComment{Comment}{/* }{ */}

\begin{algorithm}[t]
\caption{NeoCell initialization algorithm}
\KwData{Height: $h > 0$, Width: $w > 0$}
\KwResult{Initialization matrix $W_0 \in \mathbb{R}^{h \times w}$}
/* Initialize with zero matrix */ \\
$W_0 \gets 0_{h \times w}$\;
\uIf{$h < w$}{
    $step \gets round(w/h)$\;
    \For{$i \gets 0$ \KwTo $h-1$}{
        $start \gets i \cdot step$\;
        $end \gets min((i+1) \cdot step, w)$\;
        \For{$j \gets start$ \KwTo $end$}{
            $W_0[i, j] \gets \dfrac{1}{end - start}$\;
        }
    }
}
\uElseIf{$h > w$}{
    $step \gets round(h/w)$\;
    \For{$i \gets 0$ \KwTo $w-1$}{
        $start \gets i \cdot step$\;
        $end \gets min((i+1) \cdot step, h)$\;
        \For{$j \gets start$ \KwTo $end$}{
            $W_0[j, i] \gets \dfrac{1}{end - start}$\;
        }
    }
}
\Else{
    /* Square matrix, identity */ \\
    $W_0 \gets I_{h}$\;
}
/* Add Gaussian noise */ \\
$W_0 \gets W_0 + \mathcal{N}_{h \times w}(0, \dfrac{1}{\sqrt{w \cdot h}})$\;
return $W_0$
\label{alg:neo_init}
\end{algorithm}

Every NeoCell square matrix is initialized as an identity matrix. Non-square matrices are initialized as "skewed" identity matrices, i.e. behave as average pooling in the local region defined by ratio between input and output size. Additionally, Gaussian noise $\mathcal{N}(0, \dfrac{1}{\sqrt{w \cdot h}})$ is added to every element of both square and non-square matrices. Further we call the proposed initialization method "NeoInit". We investigate this method in \cref{neoinit_ablation}.

\begin{figure}
    \centering
    \begin{subfigure}[b]{0.4\textwidth}
        \includegraphics[width=\textwidth]{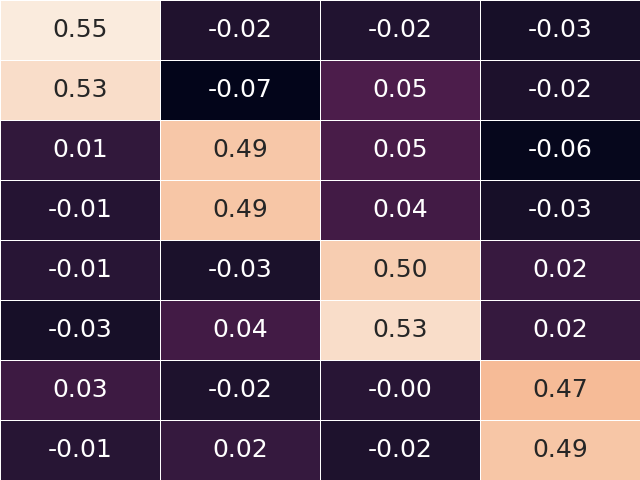}
        \caption{$8 \times 4$ matrix}
    \end{subfigure}
    \begin{subfigure}[b]{0.4\textwidth}
        \includegraphics[width=\textwidth]{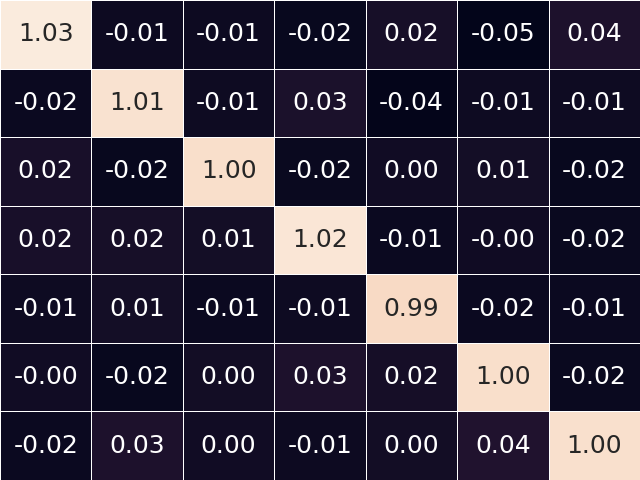}
        \caption{$7 \times 7$ matrix}
    \end{subfigure}
    \begin{subfigure}[b]{0.4\textwidth}
        \includegraphics[width=\textwidth]{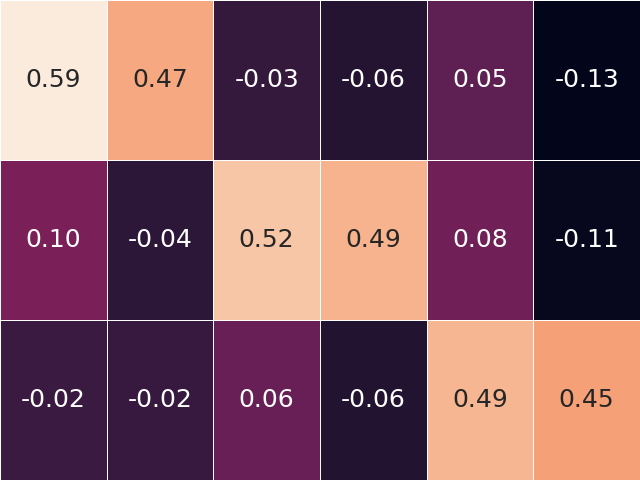}
        \caption{$3 \times 6$ matrix}
    \end{subfigure}
    \caption{Examples of NeoInit method for different matrices}
    \label{fig:neoinit_example}
\end{figure}

\section{NeoNeXt architecture}

In our architecture design we follow the scheme proposed by~\cite{liu2022convnet}, including the block structure, depth, width and resolution.
The key differences between ConvNeXt and our architecture ("NeoNeXt") are:
\begin{itemize}
    \item In the stem we use Space-To-Depth with patch size 4 which decreases width and height 4 times and increases channels number 16 times (from 3 to 48).
    Then pointwise convolution increases channels number to 96. After that Batch Normalization~\cite{https://doi.org/10.48550/arxiv.1502.03167} is applied.
    \item In model blocks we use NeoCell operator instead of the depthwise convolutions.
    In the stages 1 to 3 of the model we split all channels into two equal parts and use matrices 4x4 in the first part and 7x7 in the second part. In both parts we enable spatial shifts of the matrices (see \cref{interpatch}). In the stage 4 we use 7x7 matrices in all channels without spatial shift (as the data spatial size is already small enough and there is no need in cross-patch exchange).
    \item Instead of the Layer Normalization~\cite{https://doi.org/10.48550/arxiv.1607.06450} in NeoNeXt we prefer to use Batch Normalization~\cite{https://doi.org/10.48550/arxiv.1502.03167} because it can be efficiently fused with the previous linear operation (convolution or NeoCell) and requires no extra operations during the inference which is beneficial for the deployment.
    \item For the down-sampling between the stages and to change the width from $C_{i}$ to $C_{i+1}$ we use the following sequence of operations:\
    
    $NeoCell(h=2,h'=1,w=2,w'=1) \rightarrow $ \
    
    $ BatchNorm \rightarrow GELU \rightarrow$ \
    
    $PointwiseConv(C_{i}, C_{i+1}) \rightarrow BatchNorm \rightarrow GELU$.
\end{itemize}

\subsection{ImageNet-1K experiments} \label{imagenet_experiments}
In this section we evaluate our NeoNeXt family of models with NeoCell operation on well-known ImageNet-1K~\cite{imagenet15russakovsky} image classification benchmark.
We mostly follow the training recipe from ConvNeXt paper~\cite{liu2022convnet}. We run our models for 300 epochs, including 20 epochs of linear learning rate warmup.
We use AdamW optimizer with $\beta_{1}=0.9$ and $\beta_{2}=0.999$.
As for the regularization techniques we apply weight decay of $0.05$, label smoothing of $0.1$, randaugment~\cite{cubuk2020randaugment}, mixup~\cite{zhang2017mixup}, cutmix~\cite{yun2019cutmix} and random erasing.
We also tried exponential moving average (EMA)~\cite{polyak1992acceleration}, but it did not improve the accuracy. The details can be found in \cref{table:hyperparameters_imagenet_table}.

We have trained three models of different size, namely, NeoNeXt-T/S/B with 28M/50M/87M parameters, correspondingly, and compared them to the previously known classification models.
We show that NeoNeXt models demonstrate competitive performance (\cref{table:isynet_imagenet_table}) with the best model achieving 82.23\% top-1 accuracy at resolution 224 and 83.32\% at resolution 384.
While we do not achieve state-of-the-art performance, we prove that NeoCell operator works as intended and serves a good replacement for depth-wise convolutions.
We also show that NeoNeXt models scale both with model size and image resolution.
Moreover, NeoNext models are more efficient than ConvNeXt with respect to FLOPs according to \cref{table:isynet_imagenet_table}, but this effect is not very significant due to small share of depth-wise convs in the total number of model's FLOPs.


\begin{table}[t]
\centering
\caption{\textbf{Classification results on ImageNet-1K dataset.} We compare with several well-known models with comparable number of parameters and FLOPs. NeoNeXt does not beat SoTA, but achieves comparable results, which proves its viability.}
\begin{tabular}{l|c|c|c|c}
Model & res & \# params & GFLOPs & Top-1 \cr
\hline\hline
DeiT-S~\cite{deit}     & 224 & 22M & 4.60  & 79.8 \cr
NeoNeXt-T  & 224 & 27.7M & 4.40  & 80.87 \cr
ConvNeXt-T & 224 & 28.7M & 4.50  & 82.10 \cr
\hline
NeoNeXt-S & 224 & 49.6M & 8.60  & 82.03 \cr
ConvNeXt-S & 224 & 50.2M & 8.70  & 83.10 \cr
\hline
DeiT-B~\cite{deit} & 224 & 87M & 17.6  & 81.8 \cr
NeoNeXt-B & 224 & 87.3M & 15.19 & 82.23 \cr
ConvNeXt-B & 224 & 88.6M & 15.4  & 83.80 \cr
\hline\hline
NeoNeXt-T & 384 & 27.7M & 13.27 & 81.36 \cr
\hline
NeoNeXt-S & 384 & 49.6M & 25.71 & 83.32 \cr
\hline
NeoNeXt-B & 384 & 87.3M & 45.21 & 83.11 \cr
\hline
\end{tabular}
\label{table:isynet_imagenet_table}
\end{table}

\begin{table}[t]
\centering
\caption{ImageNet-1K training parameters}
\begin{tabular}{l|c}
                & NeoNeXt-T/S/B \cr
hyperparameters & ImageNet-1K \cr
                & $224^2$ / $384^2$ \cr
\hline
weight init     & neoinit \cr
optimizer       & AdamW  \cr
base LR         & 4e-3/4e-3/2e-3 \cr
weight decay    & 0.05  \cr
momentum        & (0.9, 0.999) \cr
batch size      & 4096/4096/2048 \cr
epochs          & 300 \cr
LR schedule     & cosine \cr
warmup epochs   & 20 \cr
warmup schedule & linear \cr
randaugment     & n=2, m=9 \cr
mixup           & 0.8 \cr
cutmix          & 1.0  \cr
random erasing  & 0.25 \cr
label smoothing & 0.1 \cr
drop path       & 0.1/0.4/0.5 \cr
gradient clip   & None \cr
EMA decay       & None \cr
\hline
\end{tabular}
\label{table:hyperparameters_imagenet_table}
\end{table}

\subsection{Ablation Study of NeoCell Initialization} \label{neoinit_ablation}

\begin{figure}[tb]
\begin{center}
\includegraphics[width=1.0\linewidth]{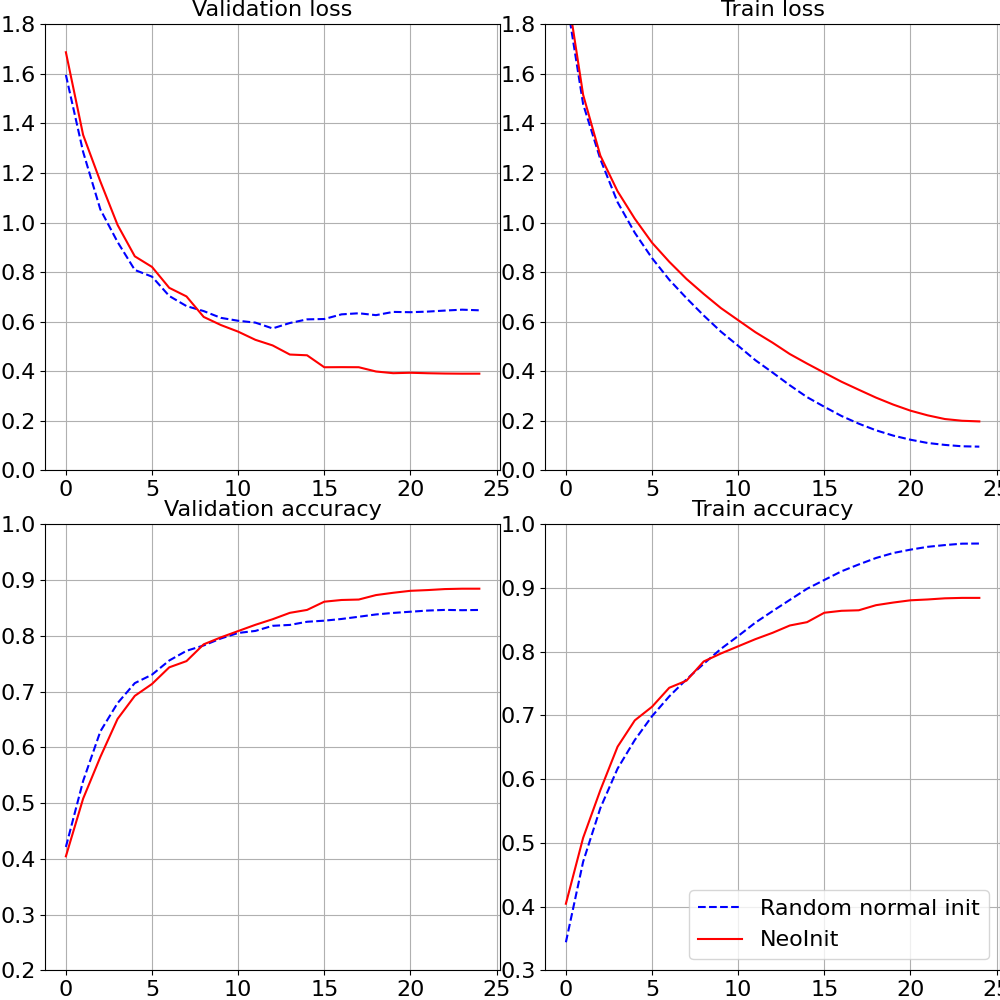}
\end{center}
   \caption{NeoInit CIFAR-10 ablation experiments}
\label{fig:neoinit_ablation}
\end{figure}

To investigate the impact of the NeoInit method we made experiments on the CIFAR-10 dataset~\cite{krizhevsky2009learning}. We train NeoNeXt-T model for 25 epochs with SGD optimizer with momentum 0.9, batch size 64. During the first epoch we use linear learning rate warmup to the maximum value 0.1, then cosine annealing schedule is applied.

\cref{fig:neoinit_ablation} and \cref{table:isynet_neoinit_ablations} show the comparison of the proposed NeoInit initialization method (see \cref{neoinit}) with the baseline random normal initialization of the NeoCell matrices (without the identity matrix): $L_{0}=\mathcal{N}(0, \dfrac{1}{\sqrt{h}})$ and $R_{0}=\mathcal{N}(0, \dfrac{1}{\sqrt{w}})$.
Each curve is averaged over 5 runs with different random seeds.

Experimental results show that the proposed initialization method allows to avoid overfitting and improve final validation accuracy of the model by 3.8 percentage points.

Interesting that with random normal initialization (i.e. without NeoInit method) approximately 75\% of the experiments diverged.
In contrast, all NeoInit experiments converged well with the same settings.
This indicates positive impact of the NeoInit method to the training convergence stability.

\section{Discussion and Future Work}

Many interesting research directions are left outside the scope of this paper.
The following directions can be considered for the future work:
\begin{enumerate}
    \item Theoretical study of the NeoCell properties, such as generalization ability, convergence, analysis of the learned matrix patterns~\etc;
    \item Design of the optimized low-level operators to better utilize hardware capabilities. We did our proof-of-concept experiments in PyTorch framework using Python API. Then we made NeoCell implementation using C++ API, which led to about 3 times acceleration. Still, we think that low-level optimization can significantly accelerate both training and inference;
    \item Search of better architecture and adaptation of the NeoCell-based architecture to the specialized NPU hardware similarly to~\cite{Tan2019efficientnet} or~\cite{letunovskiy2022isynet}. Currently, the main computational bottleneck in the current architecture is not NeoCell operators, but pointwise convolutions. Generalization to 3D case can be studied, i.e. similar to non depthwise convolution, where filter is applied to the group of or all input channels;
    \item Application of the modern training tricks to maximize the model quality, including masked autoencoders paradigm~\cite{Woo_2023_CVPR} and use of large pre-training datasets~\cite{kolesnikov2020big}.
    \item Investigation of the new applications, where NeoCell can be beneficial. This includes both classical computer vision problems such as detection or segmentation and modern multi-modal directions like text-to-image retrieval and generation of images and videos.
    \item Use of continuous weights representation instead of discrete, similar to~\cite{Solodskikh_2023_CVPR}. 
    \item Use of fast matrix multiplication algorithms to compute NeoCell operator kernel. It may allow to further  decrease the complexity and even make it sub-linear with respect to the NeoCell matrix size according to \cref{eq:neocell_complexity}.
\end{enumerate}

\section{Conclusions}

NeoCell operation can be considered as a replacement of depthwise convolutions having less computational complexity. Our experiments show that the architectures built using NeoCell show good performance on computer vision problems. Although our models don't outperform state-of-the-art, we are convinced that the proposed method has high potential due to its simplicity and flexibility and can become a foundation for the new architectures. Simple and efficient design of the proposed operation enable further improvement based on decompositions, special parameterization and well-studied linear algebra theory.

\appendix

\begin{table}[t]
\centering
\caption{NeoInit ablation study, CIFAR-10 dataset. Results are averaged over 5 runs with different random seeds.}
\begin{tabular}{|c|c|c|c|}
\hline
Model & Init method & Validation loss & Validation accuracy, \% \cr
\hline\hline
NeoNeXt-T & Random normal & 0.573 & 84.65 \cr
\hline
NeoNeXt-T & NeoInit & \textbf{0.390 (-0.183)} & \textbf{88.45 (+3.8)} \cr
\hline
\end{tabular}
\label{table:isynet_neoinit_ablations}
\end{table}

\newpage
%
%
\bibliographystyle{splncs04}
\bibliography{main}
\end{document}